\documentclass[11pt,a4paper]{article}
\usepackage[hyperref]{emnlp2020}
\usepackage{times}
\usepackage{latexsym}

\usepackage{amsmath}
\usepackage{amssymb}
\usepackage{graphicx}

\usepackage{arydshln}
\usepackage{booktabs}

\usepackage{xcolor}
\newcommand\crule[3][black]{\textcolor{#1}{\rule{#2}{#3}}}

\usepackage{microtype}

\aclfinalcopy 
\def\aclpaperid{2388} 

\title{Pushing the Limits of AMR Parsing with Self-Learning}

\author{Young-Suk Lee\thanks{\hspace{2mm}Equal contribution.},  Ram\'{o}n Fernandez Astudillo\footnotemark[1], Tahira Naseem\footnotemark[1], \\ \textbf{Revanth Gangi Reddy\footnotemark[1]\hspace{1.5mm}\thanks{\hspace{2mm}Work done during AI Residency at IBM Research.}, Radu Florian, Salim Roukos}
\\
IBM Research\\
\texttt{\{ysuklee,tnaseem\}@us.ibm.com  ramon.astudillo@ibm.com} \\}

\date{}

\begin{document}
\maketitle
\begin{abstract}
Abstract Meaning Representation (AMR) parsing has experienced a notable growth in performance in the last two years, due both to the impact of transfer learning and the development of novel architectures specific to AMR. At the same time, self-learning techniques have helped push the performance boundaries of other natural language processing applications, such as machine translation or question answering. In this paper, we explore different ways in which trained models can be applied to improve AMR parsing performance, including generation of synthetic text and AMR annotations as well as refinement of actions oracle. We show that, without any additional human annotations, these techniques improve an already performant parser and achieve state-of-the-art results on AMR 1.0 and AMR 2.0.
\end{abstract}

\section{Introduction}
\label{section:intro}
Abstract Meaning Representation (AMR) are broad-coverage sentence-level semantic representations expressing \textit{who does what to whom}. Nodes in an AMR graph correspond to concepts such as entities or predicates and are not always directly related to words. Edges in AMR represent relations between concepts such as subject/object.

AMR has experienced unprecedented performance improvements in the last two years, partly due to the rise of pre-trained transformer models \cite{radford2019language,devlin-etal-2019-bert,liu2019roberta}, but also due to AMR-specific architecture improvements. A non-exhaustive list includes latent node-word alignments through learned permutations \cite{lyu-titov-2018-amr}, minimum risk training via REINFORCE \cite{naseem-etal-2019-rewarding}, a sequence-to-graph modeling of linearized trees with copy mechanisms and re-entrance features \cite{zhang-etal-2019-amr} and more recently a highly performant graph-sequence iterative refinement model \cite{cai2020amr} and a hard-attention transition-based parser \cite{anon2020a}, both based on the Transformer architecture.

Given the strong improvements in architectures for AMR, it becomes interesting to explore alternative avenues to push performance even further. AMR annotations are relatively expensive to produce and thus typical corpora have on the order of tens of thousands of sentences. In this work we explore the use self-learning techniques as a means to escape this limitation. 

We explore the use of a trained parser to iteratively refine a rule-based AMR oracle \cite{ballesteros-al-onaizan-2017-amr,anon2020a} to yield better action sequences. We also exploit the fact that a single AMR graph maps to multiple sentences in combination with AMR-to-text \cite{mager-etal-2020-gpt-too}, to generate additional training samples without using external data. Finally we revisit silver data training \cite{konstas-etal-2017-neural}. These techniques reach $77.3$ and $80.7$ Smatch \cite{cai-knight-2013-smatch} on AMR1.0 and AMR2.0 respectively using only gold data as well as $78.2$ and $81.3$ with silver data.

\begin{figure*}[t!]
\centering
  \includegraphics[width=16cm]{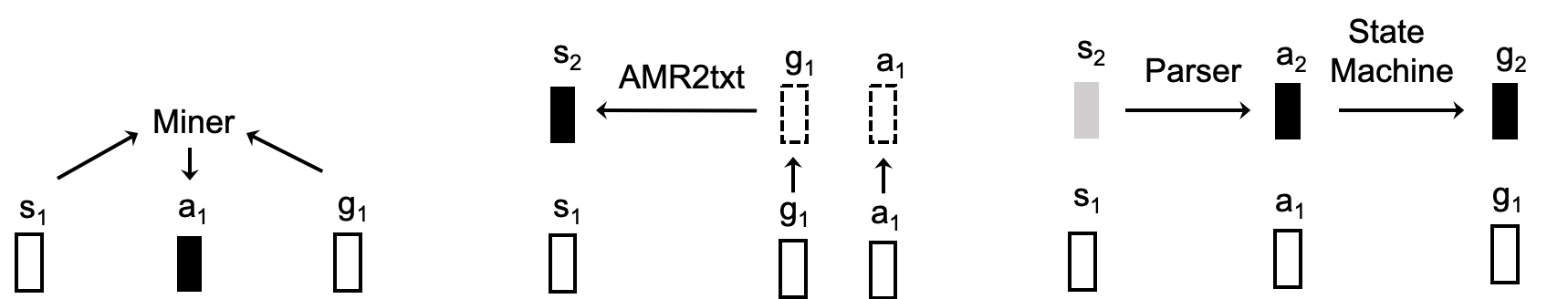}
  \caption{Role of sentence $s$, AMR graph $g$ and oracle actions $a$ in the different self-learning strategies. Left: Replacing rule-based actions by machine generated ones. Middle: synthetic text generation for existing graph annotations. Right: synthetic AMR generation for external data. Generated data (\crule{2mm}{2mm}). External data (\crule[black!30!white!100]{2mm}{2mm}). }

  \label{fig:tbp}
\end{figure*}

\section{Baseline Parser and Setup}
\label{section:baseline}
To test the proposed ideas, we used the AMR setup and parser from \cite{anon2020a} with improved embedding representations. This is a transition-based parsing approach, following the original AMR oracle in \cite{ballesteros-al-onaizan-2017-amr} and further improvements in \cite{naseem-etal-2019-rewarding}.

Briefly, rather than predicting a graph $g$ from a sentence $s$ directly, transition-based parsers predict instead an action sequence $a$. This action sequence, when applied to a state machine, produces the graph $g = M(a, s)$. This turns the problem of predicting the graph into a sequence to sequence problem, but introduces the need for an oracle to determine the action sequence $a = O(g, s)$. As in previous works, the oracle in \cite{anon2020a} is rule-based, relying on external word-to-node alignments \cite{flanigan2014discriminative,pourdamghani2016generating} to determine action sequences. It however force-aligns unaligned nodes to suitable words, notably improving oracle performance. 

As parser, \cite{anon2020a} introduces the stack-Transformer model. This is a modification of the sequence to sequence Transformer \cite{vaswani2017attention} to account for the parser state. It modifies the cross-attention mechanism dedicating two heads to attend the stack and buffer of the state machine $M(a, s)$. This parser is highly performant achieving the best results for a transition-based parser as of date and second overall for AMR2.0 and tied with the best for AMR1.0.

The stack-Transformer is trained as a conventional sequence to sequence model of $p(a \mid s)$ with a cross entropy loss. We used the full stack and full buffer setting from \cite{anon2020a} with same hyper-parameters for training and testing with the exception of the embeddings strategy detailed below. All models use checkpoint averaging \cite{junczys-dowmunt-etal-2016-amu} of the best 3 checkpoints and use a beam size of $10$\footnotemark\footnotetext{This increased scores at most $0.8$/$0.4$ for AMR1.0/2.0.} while decoding. We refer to the original paper for exact details. 

Unlike in the original work, we use RoBERTa-large, instead of RoBERTa-base embeddings, and we feed the average of all layers as input to the stack-Transformer. This considerably strengthens the baseline model from the original $76.3/79.5$ for the AMR1.0/AMR2.0 development sets to $77.6/80.8$ Smatch\footnotemark\footnotetext{We used the latest version available, $1.0.4$}. This baseline will be henceforth referred to as \cite{anon2020a} plus Strong Embeddings ($+$SE).

\section{Oracle Self-Training}
\label{section:oracle}
As explained in Section \ref{section:baseline}, transition-based parsers require an Oracle $a = O(g, s)$ to determine the action sequence producing the graph $g = M(a, s)$. Previous AMR oracles \cite{ballesteros-al-onaizan-2017-amr,naseem-etal-2019-rewarding,anon2020a} are rule based and rely on external word-to-node alignments. Rule-based oracles for AMR are sub-optimal and they do not always recover the original graph. The oracle score for AMR 2.0, measured in Smatch, is $98.1$ \cite{anon2020a} and $93.7$ for \cite{naseem-etal-2019-rewarding}. In this work, we explore the idea of using a previously trained parser, $p(a \mid s)$ to improve upon an existing oracle, initially rule-based. 

For each training sentence $s$ with graph $g$ and current oracle action sequence $a^*$, we first sample an action sequence $\tilde{a} \sim p(a \mid s)$. Both $\tilde{a}$ and $a^*$ are run through the state machine $M()$ to get graphs $\tilde{g}$ and $g^*$ respectively. We then replace $a^*$ by $\tilde{a}$ if $\mathrm{Smatch}(\tilde{g},g) > \mathrm{Smatch}(g^*,g)$ or 
$(\mathrm{Smatch}(\tilde{g},g) = \mathrm{Smatch}(g^*,g)$ and $|\tilde{a}|<|a^*|)$. This procedure is guaranteed to either increase Smatch, shorten action length or leave it unaltered. The downside is that many samples have to be drawn in order to obtain a single new best action, we therefore refer to this method as \textit{mining}. 

Starting from the improved \cite{anon2020a}, we performed $2$ rounds of mining, stopping after less than $20$ action sequences were obtained in a single epoch, which takes around $10$ epochs\footnotemark\footnotetext{One round of mining takes around $20$h, while normal model training takes $6$h on a Tesla V100.}. Between rounds we trained a new model from scratch with the new oracle to improve mining. This led to $2.0$\% actions with better Smatch and $3.7$\% shorter length for AMR1.0 and $2.8$\% and $3.2$\% respectively for AMR2.0. This results in an improvement in oracle Smatch from $98.0$ to $98.2$ for AMR 1.0 and $98.1$ to $98.3$ for AMR 2.0.

Table \ref{tab:miningovergold} shows that mining for AMR leads to an overall improvement of up to $0.2$ Smatch across the two tasks with both shorter sequences and better Smatch increasing model performance when combined. Example inspection revealed that mining corrected oracle errors such as detached nodes due to wrong alignments. It should also be noted that such type of errors are much more present in previous oracles such as \cite{naseem-etal-2019-rewarding} compared to \cite{anon2020a} and margins of improvement are therefore smaller.


\begin{table}[h]
    \centering
    \fontsize{10pt}{12pt}\selectfont
    \setlength{\tabcolsep}{2.0pt}
    \begin{tabular}{c | c | c }
        Technique & AMR1.0 & AMR2.0 \\
        \hline
        \cite{anon2020a}$+$SE & 77.6 \footnotesize{$\pm$0.1}   & 80.8 \footnotesize{$\pm$0.1}   \\
        \hline
        $<$ length $\cup$ $>$ smatch  & 77.8 \footnotesize{$\pm$0.1} & 80.9 \footnotesize{$\pm$0.2}\\
        \hline
    \end{tabular}
    \caption{Dev-set Smatch for AMR 1.0 and AMR 2.0 for different mining criteria. Average results for $3$ seeds with standard deviation.}
    \label{tab:miningovergold}
\end{table}

\section{Self-Training with Synthetic Text}
\label{section:synthetictext}
AMR abstracts away from the surface forms i.e. one AMR graph corresponds to many different valid sentences. The AMR training data, however, provides only one sentence per graph with minor exceptions. AMR 1.0 and AMR 2.0 training corpora have also only $10$k and $36$k sentences, respectively, making generalization difficult. We hypothesize that if the parser is exposed to allowable variations of text corresponding to each gold graph, it will learn to generalize better.

To this end, we utilize the recent state-of-the-art AMR-to-text system of \newcite{mager-etal-2020-gpt-too}, a generative model based on fine-tuning of GPT-2 \cite{radford2019language}. We use the trained model $p(s \mid g)$ to produce sentences from gold AMR graphs. For each graph $g$ in the training data, we generate $20$ sentences via sampling $\tilde{s} \sim p(s \mid g)$ and one using the greedy best output. We then use the following cycle-consistency criterion to filter this data. We use the improved stack-Transformer parser in Table \ref{tab:miningovergold} to generate two AMR graphs: one from the generated text $\tilde{s}$, $\tilde{g}$ and one from the original text $s$, $\hat{g}$. We then use the Smatch between these two graphs to filter out samples, selecting up to three samples per sentence if their Smatch was not less than $80.0$. We remove sentences identical to the original gold sentence or repeated. Filtering prunes roughly $90\%$ of the generated sentences. This leaves us with 18k additional sentences for AMR 1.0 and 68k for AMR 2.0. Note that the use of parsed graph, rather than the gold graph, for filtering accounts for parser error and yielded better results as a filter.

Two separate GPT-2-based AMR-to-text systems were fine-tuned using AMR 1.0 and AMR 2.0 train sets and then sampled to generate the respective text data\footnote{synTxt training takes 17h for AMR 2.0 and 5h hours for AMR 1.0 on a Tesla V100. AMR-to-text training for 15 epochs takes 4.5h on AMR 1.0 and 15h on AMR 2.0.} and conventional training was carried out over the extended dataset. As shown in Table \ref{tab:syntext}, synthetic text generation, henceforth denoted synTxt, improves parser performance over the \cite{anon2020a}$+$SE baseline for AMR2.0 and particularly for AMR1.0, possibly due to its smaller size.  

\begin{table}[h]
    \centering
    \fontsize{10pt}{12pt}\selectfont
    \setlength{\tabcolsep}{2.0pt}
    \begin{tabular}{c | c | c }
        Technique&{AMR1.0}&{AMR2.0}\\
        \hline
        \cite{anon2020a}$+$SE & 77.6 \footnotesize{$\pm$0.1} &  80.8 \footnotesize{$\pm$0.1}\\
        \hline
        synTxt     & 78.2 \footnotesize{$\pm$0.1} & 81.2 \footnotesize{$\pm$0.1}\\
        \hline
    \end{tabular}
    \caption{Dev-set Smatch for AMR 1.0 and AMR 2.0. for synthetic text. Average results for $3$ seeds with standard deviation.}
    \label{tab:syntext}
\end{table}

\section{Self-Training with Synthetic AMR}
\label{section:syntheticamr}
A trained parser can be used to parse unlabeled data and produce synthetic AMR graphs, henceforth synAMR. Although these graphs do not have the quality of human-annotated AMRs, they have been shown to improve AMR parsing performance \cite{konstas2017neural,noordbos2017amr}. The performance of prior works is however not any more comparable to current systems and it is therefore interesting to revisit this approach.




For this, we used the improved \cite{anon2020a} parser of Sec.~\ref{section:baseline} to parse unlabeled sentences from the context portion of SQuAD-2.0, comprising ~$85$k sentences and ~$2.3$m tokens, creating an initial synAMR corpus. This set is optionally filtered to reduce the training corpus size for AMR 2.0 experiments and is left unfiltered for AMR 1.0, due to its smaller size. The filtering combines two criteria. First, it is easy to detect when the transition-based system produces disconnected AMR graphs. Outputs with disconnected graphs are therefore filtered out. Second, we use a cycle-consistency criteria as in Section \ref{section:synthetictext} whereby synthetic text is generated for each synthetic AMR with \cite{mager-etal-2020-gpt-too}. For each pair of original text and generated text, the synAMR is filtered out if BLEU score \cite{papineni2002bleu} is lower than a pre-specified threshold, $5$ in our experiments. Because the AMR-to-text generation system is trained on the human-annotated AMR only, generation performance may be worse on synthetic AMR and out of domain data. Consequently we apply BLEU-based filtering only to the input texts with no out of vocabulary (OOV) tokens with respect to the original human-annotated corpus. After filtering, the synAMR data is reduced to ~$58$k sentences. 

Following prior work, we tested pre-training on synAMR only, as in ~\cite{konstas2017neural}, or on the mix of human-annotated AMR and synAMR, as in ~\cite{noordbos2017amr} and then fine-tuned on the AMR1.0 or AMR2.0 corpora. Table~\ref{tab:pretrainingcorpus} shows the results for AMR1.0 and AMR2.0 under the two pre-training options. Results show that pre-training on the mix of human-annotated AMR and synAMR works better than pre-training on synAMR only, for both AMR1.0 and AMR 2.0\footnote{human+synAMR and synAMR training take about 54h and 19h respectively for AMR2.0 and 17h and 13h respectively for AMR1.0. Fine-tuning takes 4h for AMR2.0 and 3h for AMR1.0 on a Tesla V100.}. 

\begin{table}[h]
  \centering
\scalebox{0.89}{
  \begin{tabular}{c|c|c}
    Technique & AMR1.0 & AMR2.0  \\
    \hline
    \cite{anon2020a}$+$SE & 77.6 \footnotesize{$\pm$0.1} & 80.8 \footnotesize{$\pm$0.1}\\
    \hline
    synAMR only        & 78.1\footnotesize{$\pm$0.0}  &  80.7 \footnotesize{$\pm$0.0}   \\
    human+synAMR       & 78.6\footnotesize{$\pm$0.1}  &  81.6 \footnotesize{$\pm$0.0}  \\
    \hline
  \end{tabular}
}
  \caption{Dev-set Smatch for AMR1.0 and AMR2.0. for the baseline parser and synthetic AMR training. Average results for $3$ seeds with standard deviation.}
  \label{tab:pretrainingcorpus}
\end{table}

%


\section{Detailed Analysis}
\label{section:final}
\subsection{Comparison Background}

We compare the proposed methods with recent prior art in Table \ref{table:amr-test}. Pre-trained embeddings are indicated as BERT base$^b$ and large$^B$ \cite{devlin-etal-2019-bert}, RoBERTa base$^r$ and large$^R$ \cite{liu2019roberta}. Note that RoBERTA large, particularly with layer average, can be expected to be more performant then BERT. Graph Recategorization is used in \cite{lyu2018amr,zhang2019broad} and indicated as $^G$. This is a pre-processing stage that segments text and graph to identify named entities and other relevant sub-graphs. It also removes senses and makes use of Stanford's CoreNLP to lemmatize input sentences and add POS tags. Graph recategorization also requires post-processing with Core-NLP at test time to reconstruct the graph. See \cite[Sec.~6]{zhang2019broad} for details. 

\begin{table}[!t]
\centering
\scalebox{0.89}{
\begin{tabular}{l|c|c}
Model                                        & AMR1.0  & AMR2.0 \\
\hline
\cite{lyu2018amr}$^G$                        &  73.7   & 74.4 \\  
\cite{naseem-etal-2019-rewarding}$^B$        &   -     & 75.5 \\ 
\cite{zhang2019broad} $^{B,G}$               &  71.3   & 77.0 \\
\cite{anon2020a} $^r$                        &  75.4 \footnotesize{$\pm$0.0} & 79.0 \footnotesize{$\pm$0.1}\\
\cite{cai2020amr} $^b$                       &  74.0 & 78.7 \\
\cite{cai2020amr} $^{b,G}$                   &  75.4 & 80.2 \\
\hline
\cite{anon2020a}$+$SE$^R$    & 76.9 \footnotesize{$\pm$0.1} & 80.2 \footnotesize{$\pm$0.0}\\
\hline         
oracle mining                & 76.9 \footnotesize{$\pm$0.0} & 80.3 \footnotesize{$\pm$0.1} \\
synTxt                       & 77.3 \footnotesize{$\pm$0.2}   &  80.7 \footnotesize{$\pm$0.2} \\
synAMR$^U$                   & 77.6 \footnotesize{$\pm$0.1}   & 81.0 \footnotesize{$\pm$0.1} \\
\hline
mining + synTxt              & 77.5 \footnotesize{$\pm$0.1}    & 80.4 \footnotesize{$\pm$0.0}\\
mining + synAMR$^U$          & 77.7 \footnotesize{$\pm$0.1}    & 80.9 \footnotesize{$\pm$0.0}\\
synTxt + synAMR$^U$          & 78.1 \footnotesize{$\pm$0.1} & 81.0 \footnotesize{$\pm$0.2} \\
mining + synTxt + synAMR$^U$ & \textbf{78.2 \footnotesize{$\pm$0.1}} & \textbf{81.3 \footnotesize{$\pm$0.0}}\\
\hline     
\end{tabular} 
}
\caption{Test-set Smatch for AMR1.0 and AMR2.0}
\label{table:amr-test}
\end{table}

\begin{table*}[!ht]
    \centering
    \scalebox{0.89}{
    \setlength{\tabcolsep}{2.5pt} 
    \begin{tabular}{lccccccccc} 
System & Smatch & Unlabeled & No WSD & Concepts & Named Ent. & Negations & Wikif. & Reentr. & SRL    \\\hline
\cite{cai2020amr} $^b$     & 78.7                         & 81.5 & 79.2 & 88.1 & 87.1 & 66.1 & 81.3 & 63.8 & 74.5\\
\cite{cai2020amr} $^{b,G}$ & 80.2                         & 82.8 & 80.8 & 88.1 & 81.1 & \textbf{78.9} & \textbf{86.3} & 64.6 & 74.2\\\hline
\cite{anon2020a}$+$SE$^R$  & 80.2 & 84.2 & 80.7 & 88.1 & 87.5 & 64.5 & 78.8 & 70.3 & 78.2\\\hline
oracle mining &80.3 & 84.2 & 79.0 & 87.8 & 87.7 & 65.4 & 79.0 & 70.4 & 78.2 \\
synTxt  & 80.7 & 84.6 & 81.1 & 88.5 & 88.3 & 69.8 & 78.8 & 71.1 & 79.0 \\
synAMR$^U$  & 81.0 & 85.0 & 81.5 & 88.6 & 88.5 & 65.4 & 79.0 & 71.1 & 79.0 \\\hline
mining+synTxt &80.4 & 84.5 & 80.9 & 87.9 & 87.7 & 65.8 &79.3 & 70.5 & 78.5 \\
mining+synAMR$^U$ & 80.9 & 84.9 & 81.4 &88.4 & 88.0 & 66.0 & 79.3 & 70.9 & 78.9 \\\hline
synTxt+synAMR$^U$ & 81.0 & 84.9 & 81.5 & 88.6 & 88.3 & 67.4 & 78.9 & 71.5 & 79.1 \\
mining+synTxt+synAMR$^U$ & \textbf{81.3}& \textbf{85.3} & \textbf{81.8} & \textbf{88.7} & \textbf{88.7} & 66.3 & 79.2 & \textbf{71.9} & \textbf{79.4} \\
\hline 

    \end{tabular}
    }
    \caption{Detailed scoring of the final system on AMR2.0 test sets}
    \label{table:amr-test-detail}
\end{table*}

Both \cite{naseem-etal-2019-rewarding,anon2020a} use a similar transition-based AMR oracle, but \cite{naseem-etal-2019-rewarding} uses stack-LSTM and Reinforcement Learning fine-tuning. These oracles require external alignments and a lemmatizer at train time, but only a lemmatizer at test time. It is important to underline that for the presented methods we do not use additional human annotations throughout the experiments and that the only external source of data is additional text data for synthetic AMR, which we indicate with $^U$.

\subsection{Results}

As displayed in Table~\ref{table:amr-test}, the baseline system is close to the best published system with better results for AMR1.0 ($+0.8$) and worse for AMR2.0 ($-0.5$). Transition-based systems process the sentence from left to right and model the AMR graph only indirectly through its action history and the alignments of actions to word tokens. This can be expected to generate a strong inductive bias that helps in lower resource scenarios. 

Regarding the introduced methods, mining shows close to no improvement in individual results. SynAMR provides the largest gain ($0.7/0.8$) for AMR1.0/AMR2.0 while synTxt provides close to half that gain ($0.4/0.3$). The combination of both methods also yields an improvement over their individual scores, but only for AMR1.0 with a $0.9$ improvement. Combination of mining with synTxt and synAMR hurt results, however synTxt and synAMR does improve for AMR2.0 attaining a $1.1$ improvement.

Overall, the proposed approach achieves $81.3$ Smatch in AMR2.0 combining the three methods, which is the best result obtained at the time of submission for AMR2.0, improving $1.1$ over \cite{cai2020amr}. It also obtains $78.2$ for AMR1.0, which is $2.8$ points above best previous results. Excluding silver data training, synTxt achieves $80.7$ ($+0.5$) in AMR2.0 and $77.5$ ($+2.1$) with minining in AMR1.0.

We also provide the detailed AMR analysis from \cite{damonte-etal-2017-incremental} for the best previously published system, baseline and the proposed methods in Table~\ref{table:amr-test-detail}. This analysis computes Smatch for sub-sets of AMR to loosely reflect particular sub-tasks, such as Word Sense Disambiguation (WSD), Named Entity recognition or Semantic Role Labeling (SRL). The proposed approaches and the baseline consistently outperform prior art in a majority of categories and the main observable differences seems due to differences between the transition-based and graph recategorization approaches. Wikification and negation, the only categories where the proposed methods do not outperform \cite{cai2020amr}, are handled by graph recategorization post-processing in this approach. Graph recategorization comes however at the cost of a large drop in the Name Entity category, probably due to need for graph post-processing using Core-NLP. Compared to this, transition-based approaches provide a more uniform performance across categories, and in this context the presented self-learning methods are able to improve in all categories. One aspect that merits further study, is the increase in the Negation category when using synTxt, which improves $5.4$ points, probably due to generation of additional negation examples.

\section{Related Works}
Mining for gold, introduced in Section~\ref{section:oracle}, can be related to previous works addressing oracle limitations such as dynamic oracles \cite{goldberg-nivre-2012-dynamic,ballesteros-etal-2016-training}, imitation learning \cite{goodman-etal-2016-noise} and minimum risk training \cite{naseem-etal-2019-rewarding}. All these approaches increase parser robustness to its own errors by exposing it to actions that are often inferior to the oracle sequence in score. The approach presented here seeks only the small set of sequences improving over the oracle and uses them for conventional maximum likelihood training.

Synthetic text, introduced in Section~\ref{section:synthetictext}, is related to Back-translation in Machine Translation \cite{sennrich-etal-2016-improving}. The approach presented here exploits however the fact that multiple sentences correspond to a single AMR and thus needs no external data. This is closer to recent work on question generation for question answering systems \cite{alberti-etal-2019-synthetic}, which also uses cycle consistency filtering. 

Finally, regarding synthetic AMR, discussed in Section~\ref{section:syntheticamr}, with respect to prior work \cite{konstas2017neural, noordbos2017amr} we show that synthetic AMR parsing still can yield improvements for high performance baselines, and introduce the cycle-consistency filtering.

\section{Conclusions}
\label{section:conclusions}
In this work\footnotemark\footnotetext{ \url{https://github.com/IBM/transition-amr-parser/}.}, we explored different ways in which trained models can be applied to improve AMR parsing performance via self-learning. Despite the recent strong improvements in performance through novel architectures, we show that the proposed techniques improve performance further, achieving new state-of-the-art on AMR 1.0 and AMR 2.0 tasks without the need for extra human annotations.


\begin{thebibliography}{27}
\expandafter\ifx\csname natexlab\endcsname\relax\def\natexlab#1{#1}\fi

\bibitem[{Alberti et~al.(2019)Alberti, Andor, Pitler, Devlin, and
  Collins}]{alberti-etal-2019-synthetic}
Chris Alberti, Daniel Andor, Emily Pitler, Jacob Devlin, and Michael Collins.
  2019.
\newblock Synthetic {QA} corpora generation with roundtrip consistency.
\newblock In \emph{Proceedings of the 57th Annual Meeting of the Association
  for Computational Linguistics}.

\bibitem[{Ballesteros and Al-Onaizan(2017)}]{ballesteros-al-onaizan-2017-amr}
Miguel Ballesteros and Yaser Al-Onaizan. 2017.
\newblock \href {https://doi.org/10.18653/v1/D17-1130} {{AMR} parsing using
  stack-{LSTM}s}.
\newblock In \emph{Proceedings of the 2017 Conference on Empirical Methods in
  Natural Language Processing}, pages 1269--1275, Copenhagen, Denmark.
  Association for Computational Linguistics.

\bibitem[{Ballesteros et~al.(2016)Ballesteros, Goldberg, Dyer, and
  Smith}]{ballesteros-etal-2016-training}
Miguel Ballesteros, Yoav Goldberg, Chris Dyer, and Noah~A. Smith. 2016.
\newblock \href {https://doi.org/10.18653/v1/D16-1211} {Training with
  exploration improves a greedy stack {LSTM} parser}.
\newblock In \emph{Proceedings of the 2016 Conference on Empirical Methods in
  Natural Language Processing}, pages 2005--2010, Austin, Texas. Association
  for Computational Linguistics.

\bibitem[{Cai and Lam(2020)}]{cai2020amr}
Deng Cai and Wai Lam. 2020.
\newblock \href {https://doi.org/10.18653/v1/2020.acl-main.119} {{AMR} parsing
  via graph-sequence iterative inference}.
\newblock In \emph{Proceedings of the 58th Annual Meeting of the Association
  for Computational Linguistics}, pages 1290--1301, Online. Association for
  Computational Linguistics.

\bibitem[{Cai and Knight(2013)}]{cai-knight-2013-smatch}
Shu Cai and Kevin Knight. 2013.
\newblock \href {https://www.aclweb.org/anthology/P13-2131} {{S}match: an
  evaluation metric for semantic feature structures}.
\newblock In \emph{Proceedings of the 51st Annual Meeting of the Association
  for Computational Linguistics (Volume 2: Short Papers)}, pages 748--752,
  Sofia, Bulgaria. Association for Computational Linguistics.

\bibitem[{Damonte et~al.(2017)Damonte, Cohen, and
  Satta}]{damonte-etal-2017-incremental}
Marco Damonte, Shay~B. Cohen, and Giorgio Satta. 2017.
\newblock \href {https://www.aclweb.org/anthology/E17-1051} {An incremental
  parser for {A}bstract {M}eaning {R}epresentation}.
\newblock In \emph{Proceedings of the 15th Conference of the {E}uropean Chapter
  of the Association for Computational Linguistics: Volume 1, Long Papers},
  pages 536--546, Valencia, Spain. Association for Computational Linguistics.

\bibitem[{Devlin et~al.(2019)Devlin, Chang, Lee, and
  Toutanova}]{devlin-etal-2019-bert}
Jacob Devlin, Ming-Wei Chang, Kenton Lee, and Kristina Toutanova. 2019.
\newblock \href {https://doi.org/10.18653/v1/N19-1423} {{BERT}: Pre-training of
  deep bidirectional transformers for language understanding}.
\newblock In \emph{Proceedings of the 2019 Conference of the North {A}merican
  Chapter of the Association for Computational Linguistics: Human Language
  Technologies, Volume 1 (Long and Short Papers)}, pages 4171--4186,
  Minneapolis, Minnesota. Association for Computational Linguistics.

\bibitem[{F.~A.~et~al.(2020)Fernandez~Astudillo, Ballesteros,
  Naseem, Blodget, and Florian}]{anon2020a}
Ramon Fernandez~Astudillo, Miguel Ballesteros, Tahira Naseem, Austin Blodget,
  and Radu Florian. 2020.
\newblock \href {https://openreview.net/forum?id=b36spsuUAde} {Transition-based
  parsing with stack-transformers}.
\newblock In \emph{Findings of the EMNLP2020 (to appear)}.

\bibitem[{Flanigan et~al.(2014)Flanigan, Thomson, Carbonell, Dyer, and
  Smith}]{flanigan2014discriminative}
Jeffrey Flanigan, Sam Thomson, Jaime Carbonell, Chris Dyer, and Noah~A Smith.
  2014.
\newblock A discriminative graph-based parser for the abstract meaning
  representation.
\newblock In \emph{Proceedings of the 52nd Annual Meeting of the Association
  for Computational Linguistics (Volume 1: Long Papers)}, pages 1426--1436.

\bibitem[{Goldberg and Nivre(2012)}]{goldberg-nivre-2012-dynamic}
Yoav Goldberg and Joakim Nivre. 2012.
\newblock \href {https://www.aclweb.org/anthology/C12-1059} {A dynamic oracle
  for arc-eager dependency parsing}.
\newblock In \emph{Proceedings of {COLING} 2012}, pages 959--976, Mumbai,
  India. The COLING 2012 Organizing Committee.

\bibitem[{Goodman et~al.(2016)Goodman, Vlachos, and
  Naradowsky}]{goodman-etal-2016-noise}
James Goodman, Andreas Vlachos, and Jason Naradowsky. 2016.
\newblock \href {https://doi.org/10.18653/v1/P16-1001} {Noise reduction and
  targeted exploration in imitation learning for {A}bstract {M}eaning
  {R}epresentation parsing}.
\newblock In \emph{Proceedings of the 54th Annual Meeting of the Association
  for Computational Linguistics (Volume 1: Long Papers)}, pages 1--11, Berlin,
  Germany. Association for Computational Linguistics.

\bibitem[{Junczys-Dowmunt et~al.(2016)Junczys-Dowmunt, Dwojak, and
  Sennrich}]{junczys-dowmunt-etal-2016-amu}
Marcin Junczys-Dowmunt, Tomasz Dwojak, and Rico Sennrich. 2016.
\newblock \href {https://doi.org/10.18653/v1/W16-2316} {The {AMU}-{UEDIN}
  submission to the {WMT}16 news translation task: Attention-based {NMT} models
  as feature functions in phrase-based {SMT}}.
\newblock In \emph{Proceedings of the First Conference on Machine Translation:
  Volume 2, Shared Task Papers}, pages 319--325, Berlin, Germany. Association
  for Computational Linguistics.

\bibitem[{Konstas et~al.(2017{\natexlab{a}})Konstas, Iyer, Yatskar, Choi, and
  Zettlemoyer}]{konstas-etal-2017-neural}
Ioannis Konstas, Srinivasan Iyer, Mark Yatskar, Yejin Choi, and Luke
  Zettlemoyer. 2017{\natexlab{a}}.
\newblock \href {https://doi.org/10.18653/v1/P17-1014} {Neural {AMR}:
  Sequence-to-sequence models for parsing and generation}.
\newblock In \emph{Proceedings of the 55th Annual Meeting of the Association
  for Computational Linguistics (Volume 1: Long Papers)}, pages 146--157,
  Vancouver, Canada. Association for Computational Linguistics.

\bibitem[{Konstas et~al.(2017{\natexlab{b}})Konstas, Iyer, Yatskar, Choi, and
  Zettlemoyer}]{konstas2017neural}
Ioannis Konstas, Srinivasan Iyer, Mark Yatskar, Yejin Choi, and Luke
  Zettlemoyer. 2017{\natexlab{b}}.
\newblock Neural amr: Sequence-to-sequence models for parsing and generation.
\newblock In \emph{Proceedings of the 55th Annual Meeting of the Association
  for Computational Linguistics (Volume 1: Long Papers)}, pages 146--157.

\bibitem[{Liu et~al.(2019)Liu, Ott, Goyal, Du, Joshi, Chen, Levy, Lewis,
  Zettlemoyer, and Stoyanov}]{liu2019roberta}
Yinhan Liu, Myle Ott, Naman Goyal, Jingfei Du, Mandar Joshi, Danqi Chen, Omer
  Levy, Mike Lewis, Luke Zettlemoyer, and Veselin Stoyanov. 2019.
\newblock Roberta: A robustly optimized bert pretraining approach.
\newblock \emph{arXiv preprint arXiv:1907.11692}.

\bibitem[{Lyu and Titov(2018{\natexlab{a}})}]{lyu-titov-2018-amr}
Chunchuan Lyu and Ivan Titov. 2018{\natexlab{a}}.
\newblock \href {https://doi.org/10.18653/v1/P18-1037} {{AMR} parsing as graph
  prediction with latent alignment}.
\newblock In \emph{Proceedings of the 56th Annual Meeting of the Association
  for Computational Linguistics (Volume 1: Long Papers)}, pages 397--407,
  Melbourne, Australia. Association for Computational Linguistics.

\bibitem[{Lyu and Titov(2018{\natexlab{b}})}]{lyu2018amr}
Chunchuan Lyu and Ivan Titov. 2018{\natexlab{b}}.
\newblock \href {https://doi.org/10.18653/v1/P18-1037} {{AMR} parsing as graph
  prediction with latent alignment}.
\newblock In \emph{Proceedings of the 56th Annual Meeting of the Association
  for Computational Linguistics (Volume 1: Long Papers)}, pages 397--407,
  Melbourne, Australia. Association for Computational Linguistics.

\bibitem[{Mager et~al.(2020)Mager, Fernandez~Astudillo, Naseem, Sultan, Lee,
  Florian, and Roukos}]{mager-etal-2020-gpt-too}
Manuel Mager, Ram\'on Fernandez~Astudillo, Tahira Naseem, Md~Arafat Sultan,
  Young-Suk Lee, Radu Florian, and Salim Roukos. 2020.
\newblock Gpt-too: A language-model-first approach for amr-to-text generation.
\newblock In \emph{Proceedings of the 58th Annual Meeting of the Association
  for Computational Linguistics}, Seattle, USA. Association for Computational
  Linguistics.

\bibitem[{Naseem et~al.(2019)Naseem, Shah, Wan, Florian, Roukos, and
  Ballesteros}]{naseem-etal-2019-rewarding}
Tahira Naseem, Abhishek Shah, Hui Wan, Radu Florian, Salim Roukos, and Miguel
  Ballesteros. 2019.
\newblock \href {https://doi.org/10.18653/v1/P19-1451} {Rewarding {S}match:
  Transition-based {AMR} parsing with reinforcement learning}.
\newblock In \emph{Proceedings of the 57th Annual Meeting of the Association
  for Computational Linguistics}, pages 4586--4592, Florence, Italy.
  Association for Computational Linguistics.

\bibitem[{van Noord and Bos(2017)}]{noordbos2017amr}
Rik van Noord and Johan Bos. 2017.
\newblock Neural semantic parsing by character-based translation: Experiments
  with abstract meaning representations.
\newblock \emph{arXiv preprint arXiv:1705.09980v2}.

\bibitem[{Papineni et~al.(2002)Papineni, Roukos, Ward, and
  Zhu}]{papineni2002bleu}
Kishore Papineni, Salim Roukos, Todd Ward, and Wei-Jing Zhu. 2002.
\newblock Bleu: a method for automatic evaluation of machine translation.
\newblock In \emph{Proceedings of the 40th Annual Meeting of the Association
  for Computational Linguistics}, pages 311--318.

\bibitem[{Pourdamghani et~al.(2016)Pourdamghani, Knight, and
  Hermjakob}]{pourdamghani2016generating}
Nima Pourdamghani, Kevin Knight, and Ulf Hermjakob. 2016.
\newblock Generating english from abstract meaning representations.
\newblock In \emph{Proceedings of the 9th international natural language
  generation conference}, pages 21--25.

\bibitem[{Radford et~al.(2019)Radford, Wu, Child, Luan, Amodei, and
  Sutskever}]{radford2019language}
Alec Radford, Jeffrey Wu, Rewon Child, David Luan, Dario Amodei, and Ilya
  Sutskever. 2019.
\newblock Language models are unsupervised multitask learners.
\newblock \emph{OpenAI Blog}, 1(8).

\bibitem[{Sennrich et~al.(2016)Sennrich, Haddow, and
  Birch}]{sennrich-etal-2016-improving}
Rico Sennrich, Barry Haddow, and Alexandra Birch. 2016.
\newblock \href {https://doi.org/10.18653/v1/P16-1009} {Improving neural
  machine translation models with monolingual data}.
\newblock In \emph{Proceedings of the 54th Annual Meeting of the Association
  for Computational Linguistics (Volume 1: Long Papers)}, pages 86--96, Berlin,
  Germany. Association for Computational Linguistics.

\bibitem[{Vaswani et~al.(2017)Vaswani, Shazeer, Parmar, Uszkoreit, Jones,
  Gomez, Kaiser, and Polosukhin}]{vaswani2017attention}
Ashish Vaswani, Noam Shazeer, Niki Parmar, Jakob Uszkoreit, Llion Jones,
  Aidan~N Gomez, {\L}ukasz Kaiser, and Illia Polosukhin. 2017.
\newblock Attention is all you need.
\newblock In \emph{Advances in neural information processing systems}, pages
  5998--6008.

\bibitem[{Zhang et~al.(2019{\natexlab{a}})Zhang, Ma, Duh, and
  Van~Durme}]{zhang-etal-2019-amr}
Sheng Zhang, Xutai Ma, Kevin Duh, and Benjamin Van~Durme. 2019{\natexlab{a}}.
\newblock \href {https://doi.org/10.18653/v1/P19-1009} {{AMR} parsing as
  sequence-to-graph transduction}.
\newblock In \emph{Proceedings of the 57th Annual Meeting of the Association
  for Computational Linguistics}, pages 80--94, Florence, Italy. Association
  for Computational Linguistics.

\bibitem[{Zhang et~al.(2019{\natexlab{b}})Zhang, Ma, Duh, and
  Van~Durme}]{zhang2019broad}
Sheng Zhang, Xutai Ma, Kevin Duh, and Benjamin Van~Durme. 2019{\natexlab{b}}.
\newblock \href {https://www.aclweb.org/anthology/D19-1392} {Broad-coverage
  semantic parsing as transduction}.
\newblock In \emph{Proceedings of the 2019 Conference on Empirical Methods in
  Natural Language Processing and the 9th International Joint Conference on
  Natural Language Processing (EMNLP-IJCNLP)}, pages 3784--3796, Hong Kong,
  China. Association for Computational Linguistics.

\end{thebibliography}

\end{document}